\documentclass[10pt,twocolumn,letterpaper]{article}
\usepackage{cvpr}
\usepackage{times}
\usepackage{epsfig}
\usepackage{graphicx}
\usepackage{amsmath}
\usepackage{amssymb}
\usepackage{amsmath,amssymb} 
\newcommand{\RNum}[1]{\uppercase\expandafter{\romannumeral #1\relax}} 
\usepackage{color} 
\usepackage{algorithmic}
\usepackage{algorithm}

\usepackage[breaklinks=true,bookmarks=false]{hyperref}

\cvprfinalcopy 

\def\cvprPaperID{1497} 

\ifcvprfinal\fi
\begin{document}

\title{Learning to Generate Time-Lapse Videos Using Multi-Stage \\ Dynamic Generative Adversarial Networks}
\author{Wei Xiong$^{\dagger\ddagger}$\thanks{This work was primarily done while Wei Xiong was a Research Intern with Tencent AI Lab.} \hspace{0.2in} Wenhan Luo$^\dagger$ \hspace{0.2in} Lin Ma$^\dagger$ \hspace{0.2in} Wei Liu$^\dagger$ \hspace{0.2in} Jiebo Luo$^{\ddagger\dagger}$ \\
$^\dagger$Tencent AI Lab \hspace{0.4in}  $^\ddagger$University of Rochester\\
$^\ddagger${\tt\small \{wxiong5,jluo\}@cs.rochester.edu} $^\dagger${\tt\small \{whluo.china,forest.linma\}@gmail.com} $^\dagger${\tt\small wliu@ee.columbia.edu}  }

\maketitle
\thispagestyle{empty}
\begin{abstract}
Taking a photo outside, can we predict the immediate future, \textit{e.g.}, how would the cloud move in the sky? We address this problem by presenting a generative adversarial network (GAN) based two-stage approach to generating realistic time-lapse videos of high resolution. Given the first frame, our model learns to generate long-term future frames. The first stage generates videos of realistic contents for each frame. The second stage refines the generated video from the first stage by enforcing it to be closer to real videos with regard to motion dynamics. To further encourage vivid motion in the final generated video, Gram matrix is employed to model the motion more precisely. We build a large scale time-lapse dataset, and test our approach on this new dataset. Using our model, we are able to generate realistic videos of up to $128\times 128$ resolution for 32 frames. Quantitative and qualitative experiment results demonstrate the superiority of our model over the state-of-the-art models.
\end{abstract}

\section{Introduction}
\label{sec:introduction}
Humans can often estimate fairly well what will happen in the immediate future given the current scene. However, for vision systems, predicting the future states is still a challenging task. The problem of future prediction or video synthesis has drawn more and more attention in recent years since it is critical for various kinds of applications, such as action recognition \cite{simonyan2014two}, video understanding \cite{wu2016harnessing}, and video captioning \cite{yu2016video}. The goal of video prediction in this paper is to generate realistic, long-term, and high-quality future frames given one starting frame. Achieving such a goal is difficult, as it is challenging to model the multi-modality and uncertainty in generating both the content and motion in future frames. 

\begin{figure}[tbp]
	\centerline{\includegraphics[width=1\linewidth]{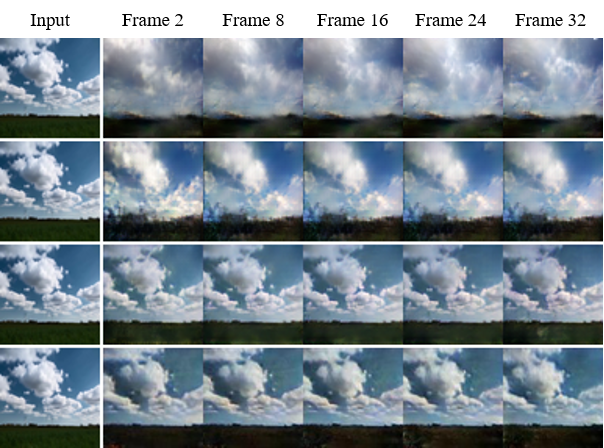}}
	\caption{From top to bottom: example frames of generated videos by VGAN \cite{vondrick2016generating}, RNN-GAN \cite{zhou2016learning}, the first stage of our model, and the second stage of our model, respectively. The contents generated by our model (the third and fourth rows) are visually more realistic. The left column is the input starting frame.}
	\label{fig.all}
    \vspace{-0.1in}
\end{figure}

In terms of content generation, the main problem is to define what to learn. Generating future on the basis of only one static image encounters inherent uncertainty of the future, which has been illustrated in \cite{vondrick2017generating}. Since there can be multiple possibilities for reasonable future scenes following the first frame, the objective function is difficult to define. Generating future frames by simply learning to reconstruct the real video can lead to unrealistic results \cite{vondrick2016generating,mathieu2015deep}. Several models including \cite{villegas2017learning} and \cite{vondrick2016generating} are proposed to address this problem based on generative adversarial networks \cite{goodfellow2014generative}. For example, 3D convolution is incorporated in an adversarial network to model the transformation from an image to a video in \cite{vondrick2016generating}. Their model produces plausible futures given the first frame. However, the generated video tends to be blurry and lose content details, which degrades the reality of generated videos. A possible cause is that the vanilla encoder-decoder structure in the generator fails to preserve all the indispensable details of the content.

Regarding motion transformation, the main challenge is to drive the given frame to transform realistically over time. Some prior work has investigated this problem. Zhou and Berg \cite{zhou2016learning} use an RNN to model the temporal transformations. They are able to generate a few types of motion patterns, but not realistic enough. The reason may be that, each future frame is based on the state of previous frames, so the error accumulates and the motion distorts over time. The information loss and error accumulation during the sequence generation hinder the success of future prediction.

The performance of the prior models indicates that it is nontrivial to generate videos with both realistic contents in each frame and vivid motion dynamics across frames with a single model at the same time. One reason may be that the representation capacity of a single model is limited in satisfying two objectives that may contradict each other. To this end, we divide the modeling of video generation into content and motion modeling, and propose a Multi-stage Dynamic Generative Adversarial Network (MD-GAN) model to produce realistic future videos. There are two stages in our approach. The first stage aims at generating future frames with content details as realistic as possible given an input frame. The second stage specifically deals with motion modeling, \textit{i.e.}, to make the movement of objects between adjacent frames more vivid, while keeping the content realistic. 

To be more specific, we develop a generative adversarial network called Base-Net to generate contents in the first stage. Both the generator and the discriminator are composed of 3D convolutions and deconvolutions to model temporal and spatial patterns. The adversarial loss of this stage encourages the generator to produce videos of similar distributions to real ones. In order to preserve more content details, we use a 3D U-net \cite{U-net} like architecture in the generator instead of the vanilla encoder-decoder structure. Skip connections \cite{he2016identity} are used to link the corresponding feature maps in the encoder and decoder so that the decoder can reuse features in the encoder, thus reducing the information loss. In this way, the model can generate better content details in each future frame, which are visually more pleasing than those produced by the vanilla encoder-decoder architecture such as the model in \cite{vondrick2016generating}. 

The Base-Net can generate frames with concrete details, but may not be capable of modeling the motion transformations across frames. To generate future frames with vivid motion, the second stage MD-GAN takes the output of the first stage as input, and refines the temporal transformation with another generative adversarial network while preserving the realistic content details, which we call Refine-Net. We propose an adversarial ranking loss to train this network so as to encourage the generated video to be closer to the real one while being further away from the input video (from stage \RNum{1}) regarding motion. To this end, we introduce the Gram matrix \cite{gatys2015texture} to model the dynamic transformations among consecutive frames. We present a few example frames generated by the conventional methods and our method in Fig. \ref{fig.all}. The image frames generated by our model are sharper than the state-of-the-art and are visually almost as realistic as the real ones. 

We build a large scale time-lapse video dataset called Sky Scene to evaluate the models for future prediction. Our dataset includes daytime, nightfall, starry sky, and aurora scenes. MD-GAN is trained on this dataset and predicts future frames given a static image of sky scene. We are able to produce $128\times128$ realistic videos, whose resolution is much higher than that of the state-of-the-art models. Unlike some prior work which generates merely one frame at a time, our model generates 32 future frames by a single pass, further preventing error accumulation and information loss.

Our key contributions are as follows:

1. We build a large scale time-lapse video dataset, which contains high-resolution dynamic videos of sky scenes.

2. We propose a Multi-stage Dynamic Generative Adversarial Network (MD-GAN), which can effectively capture the spatial and temporal transformations, thus generating realistic time-lapse future frames up to $128\times128$ resolution given only one starting frame.	

3. We introduce the Gram matrix for motion modeling and propose an adversarial ranking loss to mimic motions of real-world videos, which refines motion dynamics of preliminary outputs in the first stage and forces the model to produce more realistic and higher-quality future frames.


\section{Related Work}
\label{sec:related_work}

\noindent\textbf{Generative Adversarial Networks.}
A generative adversarial network (GAN)\cite{goodfellow2014generative,arjovsky2017wasserstein,xiao2017wasserstein,wu-cvpr-2018} is composed of a generator and a discriminator. The generator tries to fool the discriminator by producing samples similar to real ones, while the discriminator is trained to distinguish the generated samples from the real ones. GANs have been successfully applied to image generation. In the seminal paper \cite{goodfellow2014generative}, models trained on the MNIST dataset and the Toronto Face Database (TFD), respectively, generate images of digits and faces with high likelihood. Relying only on random noise, GAN cannot control the mode of the generated samples, thus conditional GAN \cite{mirza2014conditional} is proposed. Images of digits conditioned on class labels and captions conditioned on image features are generated. Many subsequent works are variants of conditional GAN, including image to image translation \cite{isola2016image, CycleGAN2017}, text to image translation \cite{reed2016generative} and  super-resolution \cite{ledig2016photo}. Our model is also a GAN conditioned on a starting image to generate a video.

Inspired by the coarse-to-fine strategy, multi-stack methods such as StackGAN \cite{zhang2016stackgan}, LAPGAN \cite{denton2015deep} have been proposed to first generate coarse images and then refine them to finer images. Our model also employs this strategy to stack GANs in two stages. However, instead of refining the pixel-level details in each frame, the second stage focuses on improving motion dynamics across frames.  

\begin{figure*}[tbp]
\vspace{-0.1in}
	\centerline{\includegraphics[width=6.3in]{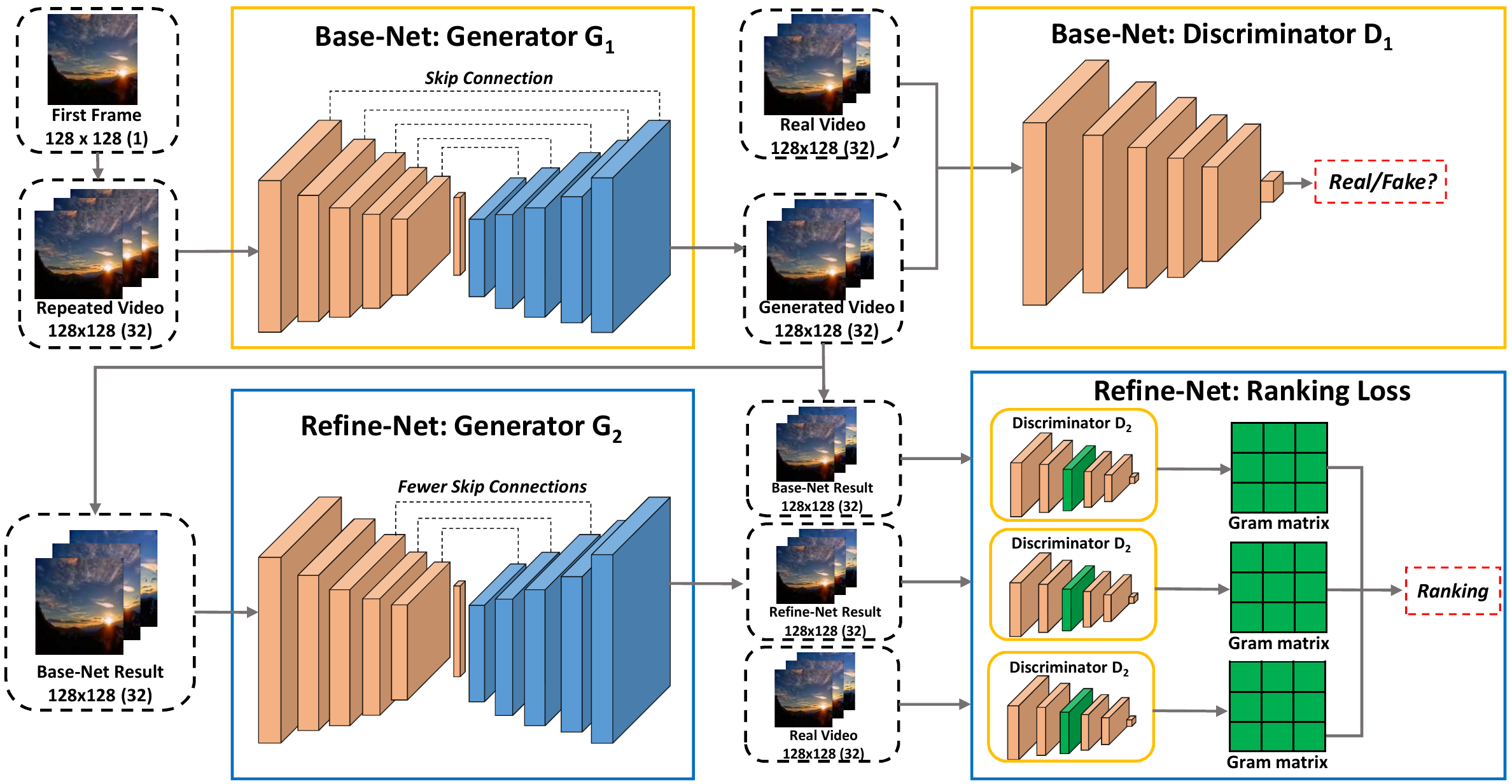}}
	\caption{The overall architecture of our MD-GAN model. The input image is first duplicated to 32 frames as input to generator $G_1$ of the Base-Net, which produces a video $\textbf{Y}_1$. Discriminator $D_1$ then distinguishes the real video $\textbf{Y}$ from $\textbf{Y}_1$. Following the Base-Net, the Refine-Net takes the generated video of $G_1$ as the input and produces a more realistic video $\textbf{Y}_2$. Discriminator $D_2$ is updated with an adversarial ranking loss to push $\textbf{Y}_2$ (the result of Refine-Net) closer to real videos.}
	\label{fig.stage1}
\end{figure*}

\noindent\textbf{Video Generation.}
Based on conditional VAE \cite{kingma2013auto}, Xue et al. \cite{visualdynamics16} propose a cross convolutional network to model layered motion, which applies learned kernels to image features encoded in a multi-scale image encoder. The output difference image is added to the current frame to produce the next frame. \cite{mathieu2015deep} is one of the earliest work that adopts generative adversarial networks to produce future frames. It uses the adversarial loss and an image gradient difference loss instead of the standard Mean Square Error to avoid blurry results. In \cite{vondrick2016generating}, a two-stream CNN, one for foreground and the other one for background, is proposed for video generation. Combining the dynamic foreground stream and the static background stream, the generated video looks real. In the follow-up work \cite{vondrick2017generating}, Vondrick and Torralba formulate the future prediction task as transforming pixels in the past to future. Based on large scale unlabeled video data, a CNN model is trained with adversarial learning. Content and motion are decomposed and encoded separately by multi-scale residual blocks, and then combined and decoded to generate plausible videos on both the KTH and the Weizmann datasets \cite{villegas2017decomposing}. A similar idea is presented in \cite{tulyakov2017mocogan}. To generate long-term future frames, Villegas et al. \cite{villegas2017learning} estimate high-level structure (human body pose), and learn a LSTM and an analogy-based encoder-decoder CNN to generate future frames based on the current frame and the estimated high-level structure. 

The closest work to ours is \cite{zhou2016learning}, which also generates time-lapse videos. However, there are important differences between their work and ours. First, our method is based on 3D convolution while a recurrent neural network is employed in \cite{zhou2016learning} to recursively generate future frames, which is prone to error accumulation. Second, as modeling motion is indispensable for video generation, we explicitly model motion by introducing the Gram matrix. Finally, we generate high-resolution ($128\times128$) videos of dynamic scenes, while the generated videos in \cite{zhou2016learning} are simple (usually with clean background) and of resolution 64$\times$64.

\section{Our Approach}
\label{sec:approach}

\subsection{Overview}

The proposed MD-GAN takes a single RGB image as input and attempts to predict future frames that are as realistic as possible. This task is accomplished in two stages in a coarse-to-fine manner: 1) Content generation by Base-Net in Stage \RNum{1}. Given an input image $\textbf{x}$, the model generates a video $\textbf{Y}_1$ of $T$ frames (including the starting frame, \textit{i.e.}, the input image). The Base-Net ensures that each produced frame in $\textbf{Y}_1$ looks like a real natural image. Besides, $\textbf{Y}_1$ also serves as a coarse estimation of the ground-truth \textbf{Y} regarding motion. 2) Motion generation by Refine-Net in Stage \RNum{2}. The Refine-Net makes efforts to refine $\textbf{Y}_1$ with vivid motion dynamics, and produces a more vivid video $\textbf{Y}_2$ as the final prediction. The discriminator $D_2$ of the Refine-Net takes three inputs, the output video $\textbf{Y}_1$ of the Base-Net, the fake video $\textbf{Y}_2$ produced by the generator of the Refine-Net and the real video \textbf{Y}. We define an adversarial ranking loss to encourage the final video $\textbf{Y}_2$ to be closer to the real video and further away from video $\textbf{Y}_1$. Note that on each stage, we follow the setting in Pix2Pix \cite{isola2016image} and do not incorporate any random noise. The overall architecture of our model is plotted in Fig. \ref{fig.stage1}.

\subsection{Stage \RNum{1}: Base-Net}
As shown in Fig. \ref{fig.stage1}, the Base-Net is a generative adversarial network composed of a generator $G_1$ and a discriminator $D_1$. Given an image $\textbf{x} \in \mathbb{R}^{3\times H \times W}$ as a starting frame, we duplicate it $T$ times, obtaining a static video $\textbf{X} \in \mathbb{R}^{3\times T \times H \times W}$ \footnote{In the generator, we can also use a 2D CNN to encode an image, but we duplicate the input image to a video to better fit our 3D U-net like architecture of $G_1$.}. By forwarding $\textbf{X}$ through layers of 3D convolutions and 3D deconvolutions, the generator $G_1$ outputs a video $\textbf{Y}_1\in \mathbb{R}^{3\times T \times H \times W}$ of $T$ frames, \textit{i.e.}, $\textbf{Y}_1 = G_1(\textbf{X})$. 

For generator $G_1$, we adopt an encoder-decoder architecture, which is also employed in \cite{radford2015unsupervised} and \cite{vondrick2016generating}. However, such a vanilla encoder-decoder architecture encounters problems in generating decent results as the features from the encoder may not be fully exploited. Therefore, we utilize a 3D U-net like architecture \cite{U-net} instead so that features in the encoder can be fully made use of to generate $\textbf{Y}_1$. This U-net architecture is implemented by introducing skip connections between the feature maps of the encoder and the decoder, as shown in Fig. \ref{fig.stage1}. The skip connections build information highways between the features in the bottom and top layers, so that features can be reused. In this way, the generated video is more likely to contain rich content details. This may seem like a simple modification, yet it plays a key role in improving the quality of videos. 

The discriminator $D_1$ then takes video $\textbf{Y}_1$ and the real video \textbf{Y} as input and tries to distinguish them. \textbf{x} is the first frame of \textbf{Y}. $D_1$ shares the same architecture as the encoder part of $G_1$, except that the final layer is a single node with a sigmoid activation function.

To train our GAN-based model, the adversarial loss of the Base-Net is defined as:
\begin{equation}
\begin{array}{l}
	\mathcal{L}_{adv} = \mathop {\min }\limits_{G_1} \mathop {\max }\limits_{D_1} {\mathbb{E}}\left[\log D_1\left(\textbf{Y}\right)\right] + \\
	\ \ \ \ \ \ \ \ \ \ \ \ \ \ \ \ \ \ \ \ \ \ \ \ \ \ \ \ \ {\mathbb{E}}\left[\log\left(1-D_1\left(G_1\left(\textbf{X}\right)\right)\right)\right] \, .
\end{array}
\end{equation}

Prior work based on conditional GAN discovers that combining the adversarial loss with an $L_1$ or $L_2$ loss \cite{isola2016image} in the pixel space will benefit the performance. Hence, we define a content loss function as a complement to the adversarial loss, to further ensure that the content of the generated video follows similar patterns to the content of real-world videos. As pointed out in \cite{isola2016image}, $L_1$ distance usually results in sharper outputs than those of $L_2$ distance. Recently, instead of measuring the similarity of images in the pixel space, perceptual loss \cite{johnson2016perceptual} is introduced in some GAN-based approaches to model the distance between high-level feature representations. These features are extracted from a well-trained CNN model and previous experiments suggest they capture semantics of visual contents \cite{ledig2016photo}. Although the perceptual loss performs well in combination with GANs \cite{ledig2016photo,li2017perceptual} on some tasks, it typically requires features to be extracted from a pretrained deep neural network, which is both time and space consuming. In addition, we observe in experiments that directly combining the adversarial loss and an $L_1$ loss that minimizes the distance between the generated video and the ground-truth video in the pixel space leads to satisfactory performance. Thus, we define our content loss as

\begin{equation}
	\mathcal{L}_{con} \left(G_1\right) =  \left\|\textbf{Y} - G_1\left(\textbf{X}\right)\right\|_1 \, .
\end{equation}

The final objective of our Base-Net in Stage \RNum{1} is
\begin{equation}
	\mathcal{L}_{stage1} =  \mathcal{L}_{adv} + \mathcal{L}_{con}\, .
\end{equation}

The adversarial training allows the Base-Net to produce videos with realistic content details. However, as the learning capacity of GAN is limited considering the uncertainty of the future, one single GAN model may not be able to capture the correct motion patterns in the real-world videos. As a consequence, the motion dynamics of the generated videos may not be realistic enough. To tackle this problem, we further process the output of Stage \RNum{1} by another GAN model called Refine-Net in Stage \RNum{2}, to compensate it for vivid motion dynamics, and generate more realistic videos.

\subsection{Stage \RNum{2}: Refine-Net}
Inputting video $\textbf{Y}_1$ from Stage \RNum{1}, our Refine-Net improves the quality of the generated video $\textbf{Y}_2$ regarding motion to fool human eyes in telling which one is real against the ground-truth video $\textbf{Y}$. 

Generator $G_2$ of the Refine-Net is similar to $G_1$ in the Base-Net. When training the model, we find it difficult to generate vivid motion while retaining realistic content details using skip connections. In other words, skip connections mainly contribute to content generation, but may not be helpful for motion generation. Thus, we remove a few skip connections from $G_2$, as illustrated in Fig. \ref{fig.stage1}.
The discriminator $D_2$ of the Refine-Net is also a CNN with 3D convolutions and shares the same structure as $D_1$ in the Base-Net. 

We adopt the adversarial training to update $G_2$ and $D_2$. However, naively employing the vanilla adversarial loss can lead to an identity mapping since the input $\textbf{Y}_1$ of $G_2$ is an optimal result of \textit{i.e.} $G_1$, which has a very similar structure as $G_2$. As long as $G_2$ learns an identity mapping, the output $\textbf{Y}_2$ would not be improved. To force the network to learn effective temporal transformations, we propose an adversarial ranking loss to drive the network to generate videos which are closer to real-world videos while further away from the input video ($\textbf{Y}_1$ from Stage \RNum{1}). The ranking loss is defined as $\mathcal{L}_{rank} \left(\textbf{Y}_1, \textbf{Y}_2, \textbf{Y}\right)$, which will be detailed later, with regard to the input $\textbf{Y}_1$, output $\textbf{Y}_2$ and the ground-truth video $\textbf{Y}$. To construct such a ranking loss, we should take the advantage of effective features that can well represent the dynamics across frames. Based on such feature representations, distances between videos can be conveniently calculated. 



We employ the Gram matrix \cite{gatys2015texture} as the motion feature representation to assist $G_2$ to learn dynamics across video frames. Given an input video, we first extract features of the video with discriminator $D_2$. Then the Gram matrix is calculated across the frames using these features such that it incorporates rich temporal information. 

Specifically, given an input video $\textbf{Y}$, suppose that the output of the $l$-\textit{th} convolutional layer in $D_2$ is $\textbf{H}_\textbf{Y}^l \in \mathbb{R}^{N \times C_l \times T_l\times H_l \times W_l}$ , where $\left(N, C_l, T_l, H_l, W_l\right)$ are the batch size, number of filters, length of the time dimension, height and width of the feature maps, respectively. We reshape $\textbf{H}_\textbf{Y}^l$ to $\hat{\textbf{H}}_\textbf{Y}^l \in \mathbb{R}^{N \times M_l \times S_l}$ , where $M_l = C_l \times T_l$ and $S_l = H_l \times W_l$. Then we calculate the Gram matrix $g(\textbf{Y};l)$ of the $n$-\textit{th} layer as follows:

\begin{equation}
g\left(\textbf{Y};l\right) = \frac{1}{M_l \times S_l} \sum\nolimits_{n=1}^N \hat{\textbf{H}}_\textbf{Y}^{l,n}\left(\hat{\textbf{H}}_\textbf{Y}^{l,n}\right)^T \, ,
\end{equation}
where $\hat{\textbf{H}}_\textbf{Y}^{l,n}$ is the $n$-\textit{th} sample of $\hat{\textbf{H}}_\textbf{Y}^l$. $g\left(\textbf{Y};l\right)$ calculates the covariance matrix between the intermediate features of discriminator $D_2$. Since the calculation incorporates information from different time steps, it can encode motion information of the given video $\textbf{Y}$.

The Gram matrix has been successfully applied to synthesizing dynamic textures in previous works \cite{Dynamictextures,twostream}, but our work differs from them in several aspects. First, we use the Gram matrix for video prediction, while the prior works use it for dynamic texture synthesis. Second, we directly calculate the Gram matrix of videos based on the features of discriminator $D_2$, which is updated in each iteration during training. In contrast, the prior works typically calculate it with a pre-trained VGG network \cite{simonyan2014very}, which is fixed during training. The motivation of such a different choice is that, as discriminator $D_2$ is closely related to the measurement of motion quality, it is reasonable to directly use features in $D_2$. 

\begin{algorithm*}[tb]
	\caption{The training procedure of the Refine-Net.}
	\label{algorithm1}
	\begin{algorithmic} 
    \STATE Set learning rates $\rho_d$ and $\rho_g$. Initialize the network parameters $\theta_d$ and $\theta_g$.
		\FOR {number of iterations}
        \STATE \textbf{Updating the discriminator $D_{2}$:}
		    
				\STATE Sample $N$ real video clips (a batch) $\{\textbf{Y}^{(1)}$, ... ,$\textbf{Y}^{(N)}\}$ from the training set.
				\STATE Obtain a batch of videos $\{\textbf{Y}_{1}^{(1)}$, ... ,$\textbf{Y}_{1}^{(N)}\}$ generated by the Base-Net.
				
		\STATE $\theta_d := \theta_d + \rho_d \nabla_{\theta_d}\dfrac{1}{N}\sum_{n=1}^N \left(\log D_{2}(\textbf{Y}^{(n)})+\log \left(1-D_{2}(G_{2}(\textbf{Y}_{1}^{(n)})) \right) + \lambda \cdot\mathcal{L}_{rank}\left(\textbf{Y}_1^{(n)},  G_{2}(\textbf{Y}_{1}^{(n)}), \textbf{Y}^{(n)} \right) \right) $ \\
		    
            \STATE \textbf{Updating the generator $G_{2}$:}
		    \STATE Sample $N$ new real video clips $\{\textbf{Y}^{(1)}$, ... ,$\textbf{Y}^{(N)}\}$ from the training set.
		    \STATE Obtain a new batch of videos $\{\textbf{Y}_{1}^{(1)}$, ... ,$\textbf{Y}_{1}^{(N)}\}$ generated by the Base-Net .
		   \STATE   $\theta_g := \theta_g -\rho_g \nabla_{\theta_g}\dfrac{1}{N}\sum_{n=1}^N \left(\log \left(1-D_{2}(G_{2}(\textbf{Y}_{1}^{(n)})) \right) + \lambda \cdot\mathcal{L}_{rank}\left(\textbf{Y}_1^{(n)},  G_{2}(\textbf{Y}_{1}^{(n)}), \textbf{Y}^{(n)}\right) + \mathcal{L}_{con}\right)  $

		\ENDFOR
	\end{algorithmic}
\end{algorithm*}

To make full use of the video representations, we adopt a variant of the contrastive loss introduced in \cite{hoffer2015deep} and \cite{contrastgan} to compute the distance between videos. Our adversarial ranking loss with respect to features from the $l$-\textit{th} layer is defined as:

\begin{equation}
\begin{array}{l}
\mathcal{L}_{rank}\left(\textbf{Y}_1, \textbf{Y}_2, \textbf{Y}; l\right) \\ = -log \dfrac{e^{-\left\|g\left(\textbf{Y}_2; l\right) - g\left(\textbf{Y}; l\right)\right\|_1}}{e^{-\left\|g\left(\textbf{Y}_2; l\right) - g\left(\textbf{Y}; l\right)\right\|_1} +e^{-\left\|g\left(\textbf{Y}_2; l\right) - g\left(\textbf{Y}_1; l\right)\right\|_1} } \, .
\end{array}
\end{equation}

We extract the features from multiple convolutional layers of the discriminator $D_2$ for the input $\textbf{Y}_1$, output $\textbf{Y}_2$ and ground-truth video $\textbf{Y}$, and calculate their Gram matrices, respectively. The final adversarial ranking loss is:

\begin{equation}
\mathcal{L}_{rank}\left(\textbf{Y}_1, \textbf{Y}_2, \textbf{Y}\right) = \sum_l \mathcal{L}_{rank}\left(\textbf{Y}_1, \textbf{Y}_2, \textbf{Y}; l\right)\, .
\end{equation}

Similar to the objective in Stage \RNum{1}, we also incorporate the pixel-wise $L_1$ distance to capture low-level details. The overall objective for the Refine-Net is:

\begin{equation}
\mathcal{L}_{stage2} = \mathcal{L}_{adv} + \lambda \cdot\mathcal{L}_{rank} + \mathcal{L}_{con}\, .
\end{equation}

As shown in Algorithm \ref{algorithm1}, the generator and discriminator are trained alternatively. When training generator $G_2$ with discriminator $D_2$ fixed, we try to minimize the adversarial ranking loss $\mathcal{L}_{rank} \left(\textbf{Y}_1, \textbf{Y}_2, \textbf{Y}\right)$, such that the distance between the generated $\textbf{Y}_2$ and the ground-truth \textbf{Y} is encouraged to be smaller, while the distance between $\textbf{Y}_2$ and $\textbf{Y}_1$ is encouraged to be larger. By doing so, the distribution of videos generated by the Refine-Net is forced to be similar to that of the real ones, and the visual quality of videos from Stage \RNum{1} can be improved.

When training discriminator $D_2$ with generator $G_2$ fixed, on the contrary, we maximize the adversarial ranking loss $\mathcal{L}_{rank} \left(\textbf{Y}_1, \textbf{Y}_2, \textbf{Y}\right)$. The insight behind is: if we update $D_2$ by always expecting that the distance between $\textbf{Y}_2$ and \textbf{Y} is not small enough, then the generator $G_2$ is encouraged to produce $\textbf{Y}_2$ that is closer to \textbf{Y} and further away from $\textbf{Y}_1$ in the next iteration. By optimizing the ranking loss in such an adversarial manner, the Refine-Net is able to learn realistic dynamic patterns and yield vivid videos.

\section{Experiments}
\label{sec:experiments}

\subsection{Dataset}
We build a relatively large-scale dataset of time-lapse videos from the Internet. We collect over 5,000 time-lapse videos from Youtube and manually cut these videos into short clips and select those containing dynamic sky scenes, such as the cloudy sky with moving clouds, and the starry sky with moving stars. Some of the clips may contain scenes that are dark or contain effects of quick zoom-in and zoom-out, thus are abandoned. 

We split the set of selected video clips into a training set and a testing set. Note that all the video clips belonging to the same long video are in the same set to ensure that the testing video clips are disjoint from those in the training set. We then decompose the short video clips into frames, and generate clips by sequentially combining continuous 32 frames as a clip. There are no overlap between two consecutive clips. We collect 35,392 training video clips, and 2,815 testing video clips, each containing 32 frames. The original size of each frame is $3\times 640 \times 360$, and we resize it into a square image of size $128\times 128$. Before feeding the clips to the model, we normalize the color values to $\left[-1, 1\right]$. No other preprocessing is required.

\begin{table*}[tbp]
	\centering
	
	\setlength\tabcolsep{0.1cm}
	\renewcommand\arraystretch{1}
	
	\caption{The architecture of the generators in both stages. The size of the input video is $3\times 32\times 128 \times 128$. }
	\label{table.architecture}
	\begin{tabular}{l|c|c|c|c|c|c|c|c|c|c|c|c}
		\hline
		Layers&conv1&conv2&conv3&conv4&conv5&conv6&deconv1&deconv2&deconv3&deconv4&deconv5&deconv6\\
		\hline
		\# Filters &32& 64&128&256&512&512&512&256&128&64&32&3 \\
		Filter Size &(3, 4, 4)&  (4, 4, 4)&  (4, 4, 4)&  (4, 4, 4)&  (4, 4, 4)&  (2, 4, 4)&  (4, 4, 4)&  (4, 4, 4)&  (4, 4, 4)&  (4, 4, 4)&  (4, 4, 4)&  (3, 4, 4)\\
		Stride      &(1, 2, 2)& (2, 2, 2) & (2, 2, 2) & (2, 2, 2) & (2, 2, 2) & (1, 1, 1) & (1, 1, 1) &  (2, 2, 2)&  (2, 2, 2)&  (2, 2, 2)&  (2, 2, 2)&  (1, 2, 2) \\
		Padding     &(1, 1, 1)& (1, 1, 1) & (1, 1, 1) & (1, 1, 1) & (1, 1, 1) & (0, 0, 0) & (0, 0, 0) &  (1, 1, 1)&  (1, 1, 1)&  (1, 1, 1)&  (1, 1, 1)&  (1, 1, 1)  \\
		\hline
	\end{tabular}
	\vspace{-0.1in}
\end{table*}

Our dataset contains videos with both complex contents and diverse motion patterns. There are various types of scenes in the dataset, including daytime, nightfall, dawn, starry night and aurora. They exhibit different kinds of foregrounds (the sky), and colors. Unlike some previous time-lapse video datasets, \textit{e.g.} \cite{zhou2016learning}, which contain relatively clean backgrounds, the backgrounds in our dataset show high-level diversity across videos. The scenes may contain trees, mountains, buildings and other static objects. It is also challenging to learn the diverse dynamic patterns within each type of scenes. The clouds in the blue sky may be of any arbitrary shape and move in any direction. In the starry night scene, the stars usually move fast along a curve in the dark sky.  

Our dataset can be used for various tasks on learning dynamic patterns, including unconditional video generation \cite{vondrick2016generating}, video prediction \cite{villegas2017learning}, video classification \cite{karpathy2014large}, and dynamic texture synthesis \cite{Dynamictextures}. In this paper, we use it for video prediction.

\subsection{Implementation Details}
The Base-Net takes a $3\times128\times128$ starting image and generates 32 image frames of resolution $128\times128$, \textit{i.e.}, $T=32$. The Refine-Net takes the output video of the Base-Net as input, and generates a more realistic video with $128\times128$ resolution. The models in both stages are optimized with stochastic gradient descent. We use Adam as the optimizer with $\beta=0.5$ and the momentum being $0.9$. The learning rate is 0.0002 and fixed throughout the training procedure.

We use Batch Normalization \cite{ioffe2015batch} followed by Leaky ReLU \cite{xu2015empirical} in all the 3D convolutional layers in both generators and discriminators, except for their first and last layers. For the deconvolutional layers, we use ReLU \cite{nair2010rectified} instead of Leaky ReLU. We use Tanh as the activation function of the output layer of the generators. The Gram matrices are calculated using the features of the first and third convolutional layers (after the ReLU layer) of discriminator $D_2$. The weight of the adversarial ranking loss is set to 1 in all experiments, \textit{i.e.}, $\lambda=1$. The detailed configurations of $G_1$ are given in Table \ref{table.architecture}. In $G_2$, we remove the skip connections between ``conv1'' and ``deconv6'', ``conv2'' and ``deconv5''. We use the identity mapping as the skip connection \cite{he2016identity}.

\begin{table}[htbp]
	\centering
	
	\renewcommand\arraystretch{1}
	
	\caption{Quantitative comparison results of different models. We show pairs of videos to a few workers, and ask them ``which is more realistic''. We count their evaluation results, which are denoted as Preference Opinion Score (POS). The value range of POS can be $\left[0, 100\right]$. If the value is greater than 50 then it means that the former performs better than the latter.}
	\label{table:quantitative}
	\begin{tabular}{l|c}
		\hline
		{``Which is more realistic?'' } & POS \\
		\hline
		Random Selection & 50 \\
		\hline
		Prefer Ours over VGAN &  92\\
		Prefer Ours over RNN-GAN & 97 \\
		\hline
		Prefer VGAN over Real & 5 \\
	  Prefer RNN-GAN over Real & 1 \\
		Prefer Ours over Real & 16\\
		\hline
	\end{tabular}
	\vspace{-0.1in}
\end{table}

\subsection{Comparison with Existing Methods}
\label{comparison_third_party}
We perform quantitative comparison between our model and the models presented in \cite{vondrick2016generating} and \cite{zhou2016learning}. For notation convenience, we name these two models as VGAN \cite{vondrick2016generating} and RNN-GAN \cite{zhou2016learning}, respectively. For a fair comparison, we reproduce the results of their models exactly according to their papers and reference codes, except some adaption to match the same experimental setting as ours. The adaption includes that, all the methods produce 32 frames as the output. Note that, both VGAN and RNN-GAN generate videos of resolution $64\times64$, so we resize the videos produced by our model to resolution $64\times64$ for fairness. 

Fig. \ref{fig.all} shows exemplar results by each method. The video frames generated by VGAN (the first row) and RNN-GAN (the second row) tend to be blurry, while our Base-Net (the third row) and Refine-Net (the fourth row) produce samples that are much more realistic, indicating that skip connections and the 3D U-net like architecture greatly benefit the content generation. 

In order to perform a more direct comparison for each model on both content and motion generation, we compare them in pairs. For each two models, we randomly select 100 clips from the testing set and take their first frames as the input. Then we produce the future prediction as a video of 32 frames by the two models. We conduct 100 times of opinion tests from professional workers based on the outputs. Each time we show a worker two videos generated from the two models given the same input frame. The worker is required to give opinion about which one is more realistic. The two videos are shown in a random order to avoid the potential issue that the worker tends to always prefer a video on the left (or right) due to laziness. Five groups of comparison are conducted in total. Apart from the comparisons between ours and VGAN and RNN-GAN, respectively, we also conduct comparisons of ours, VGAN and RNN-GAN against real videos to evaluate the performance of these models.  

Table \ref{table:quantitative} shows the quantitative comparison results. Our model outperforms VGAN \cite{vondrick2016generating} with regard to the Preference Opinion Score (POS). Qualitatively, videos generated by VGAN are usually not as sharp as ours. The following reasons are suspected to contribute to the superiority of our model. First, we adopt the U-net like structure instead of a vanilla encoder-decoder structure in VGAN. The connections between the encoder and the decoder bring more powerful representations, thus producing more concrete contents. Second, the Refine-Net makes further efforts to learn more vivid dynamic patterns. Our model also performs better than RNN-GAN \cite{zhou2016learning}. One reason may be that RNN-GAN uses an RNN to sequentially generate image frames, so their results are prone to error accumulation. Our model employs 3D convolutions instead of RNN so that the state of the next frame does not heavily depend on the state of previous frames. 

When comparing ours, VGAN and RNN-GAN with real videos, our model consistently achieves better POS than both VGAN and RNN-GAN, showing the superiority of our multi-stage model. Some results of our model are as decent as the real ones, or even perceived as more realistic than the real ones, suggesting that our model is able to generate realistic future scenes. 

\begin{figure*}[tbp]
	\centerline{\includegraphics[width=0.96\linewidth]{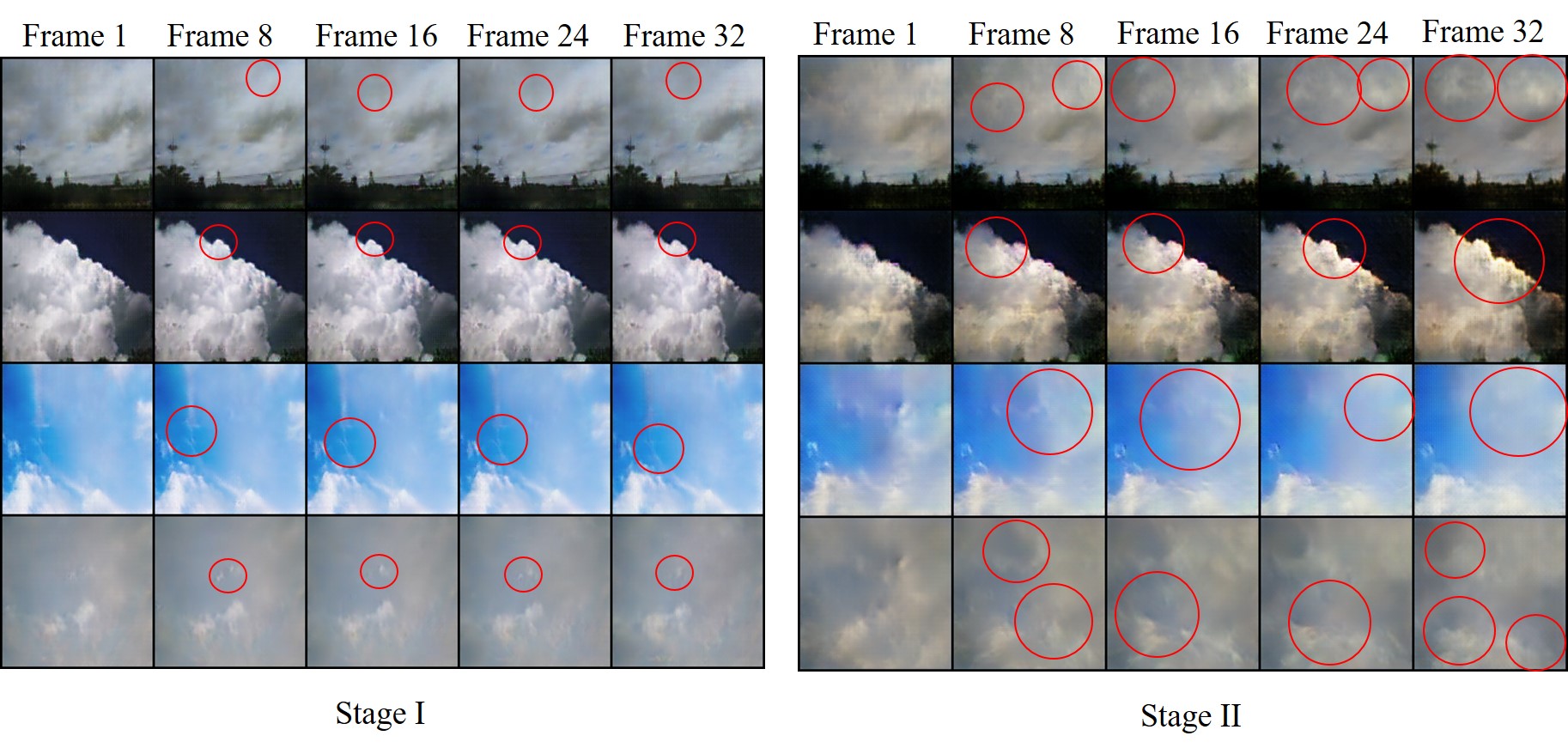}}
	\caption{The generated video frames by Stage \RNum{1} (left) and Stage \RNum{2} (right) given the same starting frame. We show exemplar frames 1, 8, 16, 24, and 32. Red circles are used to indicate the locations and areas where obvious movements take place between adjacent frames. Larger and more circles are observed in the frames of Stage \RNum{2}, indicating that there are more vivid motions generated by the Refine-Net.}
	\label{fig.selfcompare}
\end{figure*}

\subsection{Comparison between Base-Net and Refine-Net}
Although the Base-Net can generate videos of decent details and plausible motion, it fails to generate vivid dynamics. For instance, some of the results in the scene of cloudy daytime fail to exhibit apparent cloud movements. The Refine-Net makes attempts to compensate for the motion based on the result of Base-Net, while preserving the concrete content details. In this part, we evaluate the performance of Stage \RNum{2} versus Stage \RNum{1} in terms of both quantitative and qualitative results.

\begin{table}[tbp]
	\centering
	\renewcommand\arraystretch{1}
	
	\caption{Quantitative comparison results of  Stage \RNum{1} versus Stage \RNum{2}. The evaluation metric is the same as that in Table \ref{table:quantitative}.}
	\label{table:quantitative2}
	\begin{tabular}{l|c}
		\hline
		{``Which is more realistic?'' } & POS \\
		\hline
		Random Selection & 50 \\
		\hline
		Prefer Stage \RNum{2} to Stage \RNum{1} & 70 \\
		Prefer Stage \RNum{2} to Real & 16 \\	
		Prefer Stage \RNum{1} to Real & 8 \\
		\hline
	\end{tabular}
\end{table}

\noindent\textbf{Quantitative Results.} Given an identical starting frame as input, we generate two videos by the Base-Net in Stage \RNum{1} and the Refine-Net in Stage \RNum{2} separately. The comparison is carried out over 100 pairs of generated videos in a similar way to that in the previous section. Showing each pair of two videos, we ask the workers which one is more realistic. To check how effective our model is, we also compare the results of the Base-Net and Refine-Net with the ground-truth videos. The results shown in Table \ref{table:quantitative2} reveal that the Refine-Net contributes significantly to the reality of the generated videos. When comparing the Refine-Net with the Base-Net, the advantage is about 40 (70 versus 30) in terms of the POS. Not surprisingly, the Refine-Net gains better POS than the Base-Net when comparing videos of these two models with the ground-truth videos.

\noindent\textbf{Qualitative Results.}
As is shown in Fig. \ref{fig.all}, although our Refine-Net mainly focuses on improving the motion quality, it still preserves fine content details which are visually almost as realistic as the frames produced by Base-Net. In addition to content comparison, we further compare the motion dynamics of the generated video by the two stages. We show four video clips generated by the Base-Net and the Refine-Net individually on the basis of the same starting frame in Fig. \ref{fig.selfcompare}. Motions are indicated by red circles in the frames. Please note the differences between the next and previous frames. Results in Fig. \ref{fig.selfcompare} indicate that although the Base-Net can generate concrete object details, the content of the next frames seems to have no significant difference from the previous frames. While it does captures the motion patterns to some degree, like the color changes or some inconspicuous object movements, the Base-Net fails to generate vivid dynamic scene sequences. In contrast, the Refine-Net takes the output of the Base-Net to produce more realistic motion dynamics learned from the dataset. As a result, the scene sequences show more evident movements across adjacent frames. 

\subsection{Experiment on various video contexts}
Although our model works on time-lapse video generation, it can be generalized to the prediction of other video scenes. To evaluate the robustness and effectiveness of our approach, we compare our model with both VGAN and RNN-GAN on the Beach and Golf datasets released by \cite{vondrick2016generating}, which do not contain any time-lapse video. For each dataset, we use only 10\% of them as training data, and the rest as testing data. For a fair comparison, all these models take a $64\times64$ starting frame as input. To this end, we adjust our model to take $64\times64$ resolution image and video by omitting the first convolutional layer of the generators and discriminators and preserving the rest parts. For each approach, we calculate the Mean Square Error (MSE), Peak Signal to Noise Ratio (PSNR) and Structural Similarity Index (SSIM) between 1000 randomly sampled pairs of generated video and the corresponding ground-truth video. Results shown in Tables \ref{table:beach} and \ref{table:golf} demonstrate the superiority of our MD-GAN model. 

\begin{table}[tbp]
	\centering
	
	\renewcommand\arraystretch{1}
	
	\caption{Experiment results on the Beach dataset in terms of MSE, PSNR and SSIM (arrows indicating direction of better performance). The best performance values are shown in bold.}
	\label{table:beach}
	\begin{tabular}{lccc}
		\hline
		{Model } & MSE$\downarrow$  & PSNR $\uparrow$ & SSIM $\uparrow$\\
		\hline
		VGAN \cite{vondrick2016generating} & 0.0958&11.5586 &0.6035 \\
		RNN-GAN \cite{zhou2016learning}& 0.1849 &7.7988  &0.5143\\
        MD-GAN Stage \RNum{2} (Ours) & \textbf{0.0422} & \textbf{16.1951} & \textbf{0.8019}\\
		\hline
	\end{tabular}
	\vspace{-0.1in}
\end{table}

\begin{table}[tb]
\vspace{-0.02in}
	\centering
	\renewcommand\arraystretch{1}	
	\caption{Experiment results on the Golf dataset.}
	\label{table:golf}
	\begin{tabular}{lccc}
		\hline
		{Model } & MSE$\downarrow$  & PSNR $\uparrow$ & SSIM $\uparrow$\\
		\hline
		VGAN \cite{vondrick2016generating}  & 0.1188&9.9648 &0.5133 \\
		RNN-GAN \cite{zhou2016learning} & 0.2333 &7.7583  &0.4306\\
        MD-GAN Stage \RNum{2} (Ours) & \textbf{0.0681} & \textbf{13.7870} & \textbf{0.7085}\\
		\hline
	\end{tabular}
	\vspace{-0.2in}
\end{table}

\section{Conclusions}
\label{sec:conclusion}
We propose the MD-GAN model which can generate realistic time-lapse videos of resolution as high as $128\times128$ in a coarse-to-fine manner. In the first stage, our model generates sharp content details and rough motion dynamics by Base-Net with a 3D U-net like network as the generator. In the second stage, Refine-Net improves the motion quality with an adversarial ranking loss which incorporates the Gram matrix to effectively model the motion patterns. Experiments show that our model outperforms the state-of-the-art models and can generate videos which are visually as realistic as the real-world videos in many cases. 

\vspace{-0.1in}
\section{Acknowledgement}
\label{sec:thanks}
This work is supported in part by New York State through the Goergen Institute for Data Science, as well as the corporate  sponsors Snap Inc. and Cheetah Mobile. 

{\small
\bibliographystyle{ieee}
\bibliography{mdgan}
}

\end{document}